\documentclass[letterpaper]{article} 
\usepackage{aaai2026}  
\usepackage{times}  
\usepackage{helvet}  
\usepackage{courier}  
\usepackage[hyphens]{url}  
\usepackage{graphicx} 
\usepackage{natbib}  
\usepackage{caption} 
\usepackage{algorithm}
\usepackage{algorithmic}

\usepackage{booktabs}
\usepackage{threeparttable}
\usepackage{multirow}

\usepackage{newfloat}
\usepackage{listings}
\DeclareCaptionStyle{ruled}{labelfont=normalfont,labelsep=colon,strut=off} 
\floatstyle{ruled}
\newfloat{listing}{tb}{lst}{}
\floatname{listing}{Listing}
\title{DeepRWCap: Neural-Guided Random-Walk Capacitance Solver for IC Design}

\author{
    Hector Rodriguez Rodriguez\equalcontrib,
    Jiechen Huang\equalcontrib,
    Wenjian Yu
}
\affiliations{
    Department of Computer Science and Technology, BNRist, State Key Laboratory of Cryptography and Digital Economy Security, Tsinghua University, Beijing 100084, China\\
    \{lad24,hjc22\}@mails.tsinghua.edu.cn, yu-wj@tsinghua.edu.cn
}

\usepackage{amssymb}
\usepackage{amsmath}
\newcommand{\rr}{\mathbf{r}}
\newcommand{\dd}{\mathrm{d}}

\newcommand{\XX}{\mathbf{X}}

\newcommand{\EE}{\mathbb{E}}

\begin{document}

\maketitle

\begin{abstract}

Monte Carlo random walk methods are widely used in capacitance extraction for their mesh-free formulation and inherent parallelism. However, modern semiconductor technologies with densely packed structures present significant challenges in unbiasedly sampling transition domains in walk steps with multiple high-contrast dielectric materials. We present DeepRWCap, a machine learning-guided random walk solver that predicts the transition quantities required to guide each step of the walk. These include Poisson kernels, gradient kernels, signs and magnitudes of weights. DeepRWCap employs a two-stage neural architecture that decomposes structured outputs into face-wise distributions and spatial kernels on cube faces. It uses 3D convolutional networks to capture volumetric dielectric interactions and 2D depthwise separable convolutions to model localized kernel behavior. The design incorporates grid-based positional encodings and structural design choices informed by cube symmetries to reduce learning redundancy and improve generalization. Trained on 100,000 procedurally generated dielectric configurations, DeepRWCap achieves a mean relative error of $1.24\pm0.53$\% when benchmarked against the commercial Raphael solver on the self-capacitance estimation of 10 industrial designs spanning 12 to 55 nm nodes. Compared to the state-of-the-art stochastic difference method Microwalk, DeepRWCap achieves an average 23\% speedup. On complex designs with runtimes over 10 s, it reaches an average 49\% acceleration.

\end{abstract}

\begin{links}
    \link{Code}{https://github.com/THU-numbda/deepRWCap}
\end{links}

\section{Introduction}

\begin{figure*}[ht]
    \centering
    \includegraphics[width=1\textwidth]{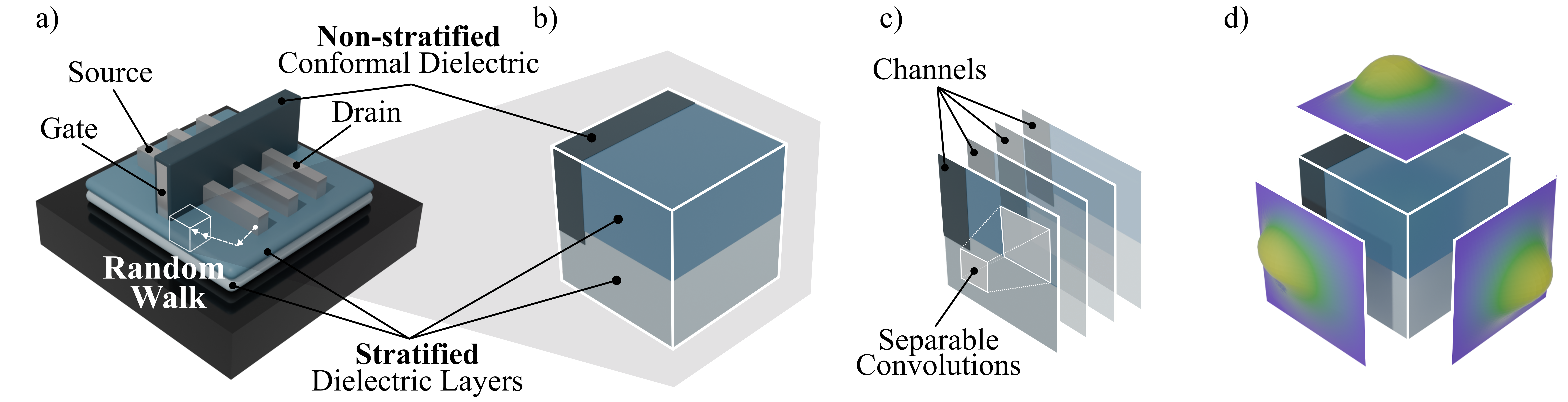}
    \caption{Illustration of the neural-guided random walk capacitance extraction framework: (a) FinFET structure, (b) transition cube, (c) compact CNN, and (d) predicted kernels.}
    \label{fig:overview}
\end{figure*}

The extraction of parasitic capacitance is a key step in the design and verification of Integrated Circuits (ICs), which involves analyzing the physical layout before fabrication to ensure that it meets timing, power consumption, and signal integrity requirements. Accurate capacitance extraction (typically within 5\% error) has become increasingly challenging as the semiconductor industry shifts toward more complex process technologies~\cite{yu2021advancements}. 
With the deceleration of Moore's Law, the industry has shifted from 2D to 3D integration, leveraging tightly packed architectures to sustain performance improvements. Transistor structures evolve from FinFETs to Gate-All-Around and Complementary Field-Effect Transistors~\cite{mukesh2022review, liebmann2021cfet}. At the circuit level, vertically stacked ICs enable smaller footprints while reducing interconnect lengths~\cite{gomes2022ponte, wuu2022vcache}. 

These increasingly complex 3D structures pose significant computational challenges for capacitance extraction. Traditional methods for capacitance extraction struggle to balance computational accuracy and efficiency in modern IC design. Electrostatic field solvers such as the finite-difference method (FDM) deliver high accuracy but suffer from poor scalability when applied to large structures~\cite{yu2013rwcap}. Pattern matching-based approaches offer computational efficiency and scalability, but provide no accuracy guarantees and rely heavily on specialized domain expertise~\cite{yang2022cnncap}.

Random walk methods provide a scalable stochastic framework for solving elliptic partial differential equations (PDEs) such as Poisson and diffusion equations. Originating from the classical Walk-on-Spheres (WoS) algorithm~\cite{muller1956some}, they leverage the duality between PDEs and stochastic processes. They have gained renewed interest in computer graphics and deep learning applications~\cite{sawhney2020monte, li2023neural, miller2024differential, nam2024poisson} due to the inherent parallelism and controllable accuracy. However, while WoS has been extended to support varying diffusion coefficients~\cite{sawhney2022grid}, it requires the coefficient to be twice-differentiable and is inapplicable to the piecewise constant dielectrics in ICs. 
Thus, the random walk method for capacitance extraction still requires computationally expensive transition evaluations to guide each step when dealing with complex multi-dielectric domains~\cite{visvardis2023deep,huang2024enhancing, huang2025efficientfrwtransitionsstochastic}.

Recent advances in machine learning have introduced AI-based techniques across various Electronic Design Automation (EDA) applications, including code generation, placement, and routing optimization~\cite{cheng2022policy, zhong2024preroutgnn, lai2025analogcoder}. In the domain of capacitance extraction, learning-based methods either replace traditional solvers by directly predicting capacitance from layout geometry~\cite{yang2022cnncap, liu2024gnncap, cai2024pctcap} or assist core computational steps in numerical algorithms such as random walk methods~\cite{visvardis2023deep}. However, the fully learning-based methods struggle to generalize across different designs and technology nodes, and the hybrid methods do not offer a compelling efficiency-accuracy trade-off.

In this work, we present DeepRWCap, which builds upon a neural-guided random walk framework as illustrated in Fig.~\ref{fig:overview}. The framework leverages compact CNNs to accelerate the random walk-based capacitance extraction. In Fig.~\ref{fig:overview}, starting from a FinFET structure embedded in heterogeneous dielectrics (Fig. \ref{fig:overview}a), a local transition cube (Fig. \ref{fig:overview}b) is extracted to represent the dielectric environment surrounding each walk step. This cube is processed by compact, depthwise-separable CNNs (Fig. \ref{fig:overview}c) that estimate one of the transition quantities, the Poisson kernel (Fig. \ref{fig:overview}d) to inform the next move in the walk. Our contributions are threefold:
(1) a unified neural architecture for predicting both Poisson and gradient kernels across multi-dielectric domains,
(2) a GPU-accelerated inference engine with producer-consumer scheduling for high-throughput sampling, and
(3) an extensive evaluation on 10 industrial test cases with different technology nodes, showing an average 23\% speedup over the state-of-the-art method~\cite{huang2025efficientfrwtransitionsstochastic} while preserving high accuracy with a mean relative error of 1.24\%.

\section{Related Work}\label{sec:2}

\paragraph{Monte Carlo Methods for Capacitance Extraction}
The Floating Random Walk (FRW) method~\cite{le1992stochastic, yu2013rwcap} adapts Monte Carlo techniques to rectilinear IC geometries using cubic transition domains. Several approaches address the computational challenge of transition kernel evaluation: OCT~\cite{yang2020floating} approximates dielectric configurations through volume-weighted averaging but introduces errors in high-contrast scenarios; AGF~\cite{huang2024enhancing} provides exact solutions for stratified dielectrics but requires cube shrinking or volume-weighted averaging to handle non-stratified cases, increasing the Monte Carlo variance or introducing approximation errors; Microwalk~\cite{huang2025efficientfrwtransitionsstochastic} achieves unbiased FRW transitions but lacks support for the first step of each walk (gradient sampling), requiring fallbacks to AGF and cube shrinking.

\paragraph{Learning-Based Methods for Capacitance Extraction.} CNN-Cap~\cite{yang2022cnncap} employs ResNet architectures to directly predict layout capacitances by representing conductors as binary masks across different layers. GNN-Cap~\cite{liu2024gnncap} uses Graph Convolutional Networks with spatial message passing to extract capacitances, achieving comparable accuracy with better scalability than CNN-based approaches. 
While these methods offer speedups over electrostatic solvers, they require retraining for each process node since dielectric distributions are learned implicitly rather than explicitly modeled.
PCT-Cap~\cite{cai2024pctcap} uses Point Cloud Transformers with 8-dimensional feature vectors that include spatial coordinates, normal vectors, and relative dielectric permittivities. While achieving higher accuracy than CNN-based methods, it approximates multi-dielectric scenarios through local averaging around each point, which inadequately captures the global dielectric structure between conductors, limiting its applicability to conformal dielectrics.

\paragraph{Hybrid Methods for Capacitance Extraction.} Visvardis et al. present deep learning-driven random walks using Group-Equivariant Convolutional Neural Networks (GE-CNN)~\cite{worrall2018cubenet} with autoencoder compression and Gaussian Mixture Models (GMM) for sampling~\cite{visvardis2023deep}. However, several limitations hinder practical deployment: the group equivariant representation creates information bottlenecks by aggressively reducing the spatial dimensions, the synchronous walker architecture underutilizes parallelism, and the CPU-only implementation results in runtimes that are 12$\times$ slower than vanilla FRW methods~\cite{visvardis2023deep}. Additionally, the restricted dielectric range $[1,10]$ and GMM's inability to capture sharp discontinuities from high-contrast interfaces limit applicability to advanced semiconductor technologies.

\section{Preliminaries}

\subsection{Capacitance Extraction}

For a system of $N_c$ conductors in a three-dimensional, multi-dielectric domain $\Omega\subset \mathbb{R}^3$, the \textit{capacitance matrix} $\mathbf{C}\in\mathbb{R}^{N_c\times N_c}$ describes the linear relation between the electric potentials $\mathbf{u}\in\mathbb{R}^{N_c}$ and the conductor charges $\mathbf{q}\in\mathbb{R}^{N_c}$~\cite{Ivica2021}:
\begin{equation}
    \mathbf{q} = \mathbf{C}\mathbf{u}.
\end{equation}
The diagonal element $C_{ii}$ is the \textit{self-capacitance} of conductor $i$, and $C_{ij}$ for $i\ne j$ is the \textit{coupling capacitance} between conductors $i$ and $j$. 

To compute $C_{ij}$, we set up an electrostatic problem where conductor $j$ is held at a unit potential ($u_j=1$) and all other conductors are grounded ($u_k=0, \forall k\ne j$). 
The target capacitance is then equal to the charge on conductor $i$, which can be expressed via Gauss's law~\cite{Griffiths_2023} as
\begin{equation}\label{eq:gauss}
    C_{ij} = q_i = -\oint_{G_i} \alpha(\rr) \frac{\partial u(\rr)}{\partial \mathbf{n}_{\rr}} \dd s,
\end{equation}
where $G_i$ is a Gaussian surface enclosing conductor $i$, $\alpha$ is the spatially varying dielectric permittivity (i.e., the diffusion coefficient in Laplace's equation), and $\mathbf{n}_{\rr}$ is the unit outward normal vector at $\rr$. The electrostatic field $u(\rr)$ satisfies the following boundary value problem of Laplace's equation:

\begin{equation}\label{eq:laplace}
    \begin{cases}
        \nabla\cdot (\alpha(\rr) \nabla u(\rr)) = 0, & \rr \in \Omega, \\
        \frac{\partial u(\rr)}{\partial \mathbf{n}_{\rr}} = 0, & \rr \in \partial\Omega_0,\\
        u(\rr) = u_k, & \rr \in \partial\Omega_k, 1\le k\le N_c,
    \end{cases}
\end{equation}
where the Neumann boundary $\partial\Omega_0$ represents the extraction window and the Dirichlet boundaries $\partial\Omega_k$ are the conductor surfaces, such that $\cup_{k=0}^{N_c} \partial\Omega_k = \partial\Omega$ and $\partial\Omega_{k_1} \cap \partial\Omega_{k_2}= \emptyset, \forall k_1\ne k_2$. 
In essence, capacitance extraction is applying numerical methods to solve \eqref{eq:laplace} and evaluate the surface integral in \eqref{eq:gauss}.

\subsection{Floating Random Walk Method}\label{sec:frw}
The connection between PDEs and stochastic processes is well established through Kakutani's theorem~\cite{kakutani1944143} and, more generally, Feynman-Kac formula~\cite{oksendal2010stochastic}. 
Specifically, the solution to Laplace's equation has the following stochastic representation
\begin{equation}\label{eq:mean}
    u(\rr) = \EE[u(\mathbf{W}_{\tau}^{\xi}) | \xi = \rr ],
\end{equation}
where $\mathbf{W}_t^{\xi}$ is a standard Brownian motion starting at $\xi$, and $\tau$ is the first hitting time of $\mathbf{W}_t^{\xi}$ to the Dirichlet boundary. 

To adapt to the rectilinear geometries of circuit layouts, the floating random walk (FRW) method uses cubes to ``hop'' in the problem domain~\cite{le1992stochastic}. The resulting FRW process forms a discrete-time Markov chain $\XX_k$, where each transition is sampled on the surface of the largest cube centered at $\XX_k$ (the so-called \textit{transition cube}). The transition probability kernel is the \textit{Poisson kernel} $p_{\alpha}$~\cite{axler2001harmonic}\footnote{It is also referred to as ``surface Green's function'' in literature~\cite{visvardis2023deep,huang2024enhancing}.}. It is the normal derivative of the Green's function for Laplace's equation in the cube and depends on the dielectric configuration $\alpha$ within the cube. Based on the FRW process, the solution to \eqref{eq:laplace} is expressed as
\begin{equation}\label{eq:poisson}
    \begin{aligned}
    u(\rr_k)&=\EE[u(\XX_{k+1}) | \XX_k=\rr_k] \\
    &= \oint_{\partial S(\rr_k)} u(\rr_{k+1}) p_{\alpha}(\rr_{k+1}| \rr_k) \dd s,
    \end{aligned}
\end{equation}
where $S(\rr_k)$ is the transition cube centered at $\rr_k$.
We recursively apply Monte Carlo integration based on \eqref{eq:poisson} until hitting the Dirichlet boundary at some $\rr_T$, in which case the boundary condition $u(\rr_T)$ is an unbiased estimator for $u(\rr_k)$. Combining this estimator and our target integral in \eqref{eq:gauss}, we obtain the FRW estimator for capacitance
\begin{equation}\label{eq:estimator}
    \hat{q_i}=-\frac{\alpha(\rr)}{f(\rr)}w_{\alpha}(\rr)s_{\alpha}(\rr,\rr_1)u(\rr_T),
\end{equation}
where $f(\rr)$ is the probability density on $G_i$, $w_{\alpha}(\rr)=\oint_{\partial S(\rr)}\left|\frac{\partial p_{\alpha} (\rr_1|\rr)}{\partial \mathbf{n}_\rr}\right| \mathrm{d} s_1$, and $s_\alpha(\rr, \rr_1)=\text{sign}(\frac{\partial p_{\alpha} (\rr_1|\rr)}{\partial \mathbf{n}_\rr})$. The transition kernel is $g_\alpha(\rr_1|\rr)=\left|\frac{\partial p_{\alpha} (\rr_1|\rr)}{\partial \mathbf{n}_\rr}\right| \big/ w_{\alpha}(\rr)$ for the first transition and $p_\alpha(\rr_{k+1}|\rr_k)$ for the subsequent transitions~\cite{yu2013rwcap}. See~\cite{huang2024floating} for the derivation and proof.

\begin{figure}
    \centering
    \includegraphics[width=\linewidth]{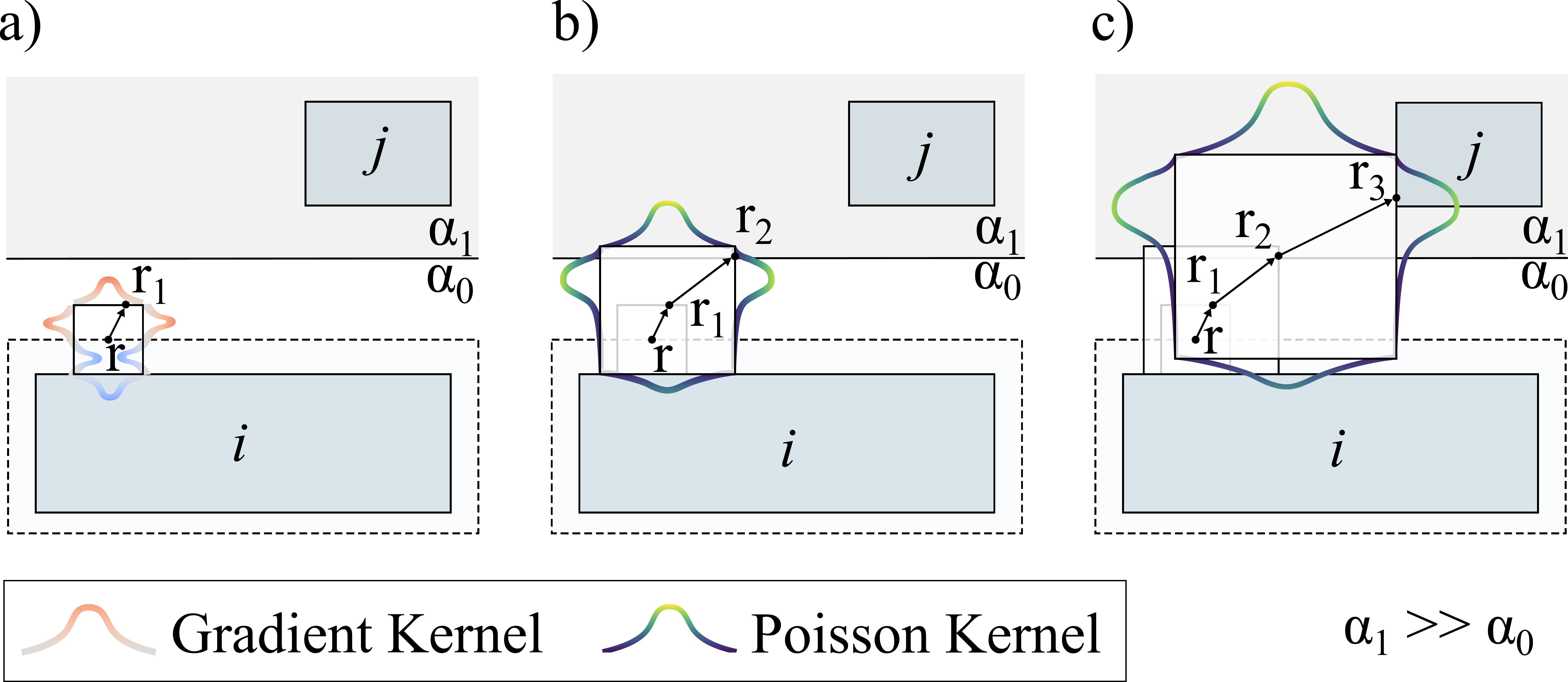}
    \caption{An FRW trajectory from conductor $i$ to $j$ with example transition kernels, showing (a) the initial step, (b) an intermediate step, and (c) the boundary hit.}
    \label{fig:frw_process}
\end{figure}

\section{DeepRWCap: Neural-Guided Random Walk Solver}

DeepRWCap is a neural-guided framework that accelerates random walk-based capacitance extraction by learning to predict transition kernels in non-stratified domains. We formulate this as a supervised learning problem and describe our synthetic dataset generation approach for training on procedurally generated dielectric configurations. We present a neural architecture that exploits cube symmetries to avoid learning redundancy and integrates with a high-throughput inference engine for efficient random walk sampling.

\subsection{Problem Formulation: Multi-Dielectric Transition Kernel Learning}
For a transition cube $S\subset \mathbb{R}^3$, let $\alpha: S\rightarrow \mathbb{R}_{>0}$ denote the relative permittivity, which is a piecewise constant function in actual IC technologies, representing the planar dielectric layers and layout-dependent dielectric blocks (see Fig. \ref{fig:overview}). 
Performing an unbiased FRW transition requires a set of transition quantities: the \textit{Poisson kernel} $p_\alpha:\partial S\rightarrow \mathbb{R}_{>0}$, the \textit{weight value} $w_\alpha\in\mathbb{R}_{>0}$, the \textit{sign distribution} $s_\alpha:\partial S\rightarrow \{-1,0,1\}$, and the \textit{gradient kernel} $g_\alpha:\partial S\rightarrow \mathbb{R}_{\ge0}$. Note that they depend only on the relative dielectric configuration inside $S$ and are independent of the absolute location of the cube. Therefore, the learning problem is to approximate the mapping from $\alpha$ to $w_\alpha,s_\alpha, g_\alpha$ for the first transition, and $p_\alpha$ for the subsequent transitions.

We discretize the transition cube on an $N\times N\times N$ voxel grid, where the dielectric $\alpha$ is represented as a tensor $\mathcal{X}\in\mathbb{R}^{N\times N\times N}$. 
Since the target quantities are invariant to global scaling of $\alpha$, we normalize each input such that $\mathcal{X}_{i,j,k}=\frac{\alpha_{i,j,k}}{\max(\alpha)} \in (0, 1]$. The target surface functions are discretized over the six cube faces as tensors $\mathbf{s}_\alpha, \mathbf{g}_\alpha, \mathbf{p}_\alpha \in \mathbb{R}^{6\times N\times N}$. 
This not only provides a robust and high-fidelity data representation, but also aligns with the standard finite-difference scheme in numerical solutions of the Poisson kernel~\cite{huang2025efficientfrwtransitionsstochastic}.

\subsection{Training Dataset Generation}
The FRW transition quantities exhibit a highly non-linear dependence on the dielectric configuration~\cite{huang2024enhancing} and the governing equations are not available in explicit forms. 
This precludes the use of physics-informed objectives~\cite{raissi2019physics} and motivates a fully data-driven learning framework. 

To enable models that generalize across different semiconductor technologies, we synthesize a dataset $\mathcal{D}=\{w_\alpha^{(i)}, \mathbf{s}_\alpha^{(i)}, \mathbf{g}_\alpha^{(i)}, \mathbf{p}_\alpha^{(i)};\mathcal{X}^{(i)} \}$ using a block-based generative procedure (Alg.~\ref{alg:dielectric}) that emulates real-world dielectric patterns. Each block is parameterized by its center $(c_x, c_y, c_z)$, side lengths $(l_x, l_y, l_z)$, and relative permittivity $\kappa$. 
To mimic materials in practical technologies, we empirically design $\mathsf{RandomDielectric}$ in Alg.~\ref{alg:dielectric} as sampling a low-$\kappa$ distribution $U(2,10)$ with $80\%$ probability or a high-$\kappa$ distribution $U(10, 80)$ with $20\%$ probability.
Nested structures are randomly added (line 6-11 in Alg.~\ref{alg:dielectric}) to mimic conformal coatings. 
Blocks in the list can overlap, with earlier ones overriding later ones. This algorithm produces high-contrast overlaps and sharper kernels compared to real layout dielectrics.

After placing the blocks via Alg.~\ref{alg:dielectric}, the domain $[-1,1]^3$ is treated as the transition cube and voxelized as the tensor representation $\mathcal{X}^{(i)}$. With a given $\mathcal{X}^{(i)}$, we use a finite-difference-method (FDM) solver~\cite{huang2025efficientfrwtransitionsstochastic} to obtain the targets $w_\alpha^{(i)}, \mathbf{s}_\alpha^{(i)}, \mathbf{g}_\alpha^{(i)}, \mathbf{p}_\alpha^{(i)}$.

\begin{algorithm}[ht]
\caption{Random Dielectric Configuration}
\label{alg:dielectric}
\textbf{Input}: Block count $B$, nesting probability $p_{\text{nest}}$\\
\textbf{Output}: A list of dielectrics $L$
\begin{algorithmic}[1]
\STATE $L \gets [\,]$
\FOR{$i=1$ \TO $B$}
    \STATE Sample $c_x,c_y,c_z \sim U(-2,2)$; $l_x,l_y,l_z \sim U(0,4)$;
    \STATE Sample $\kappa \sim \mathsf{RandomDielectric}$;
    \STATE Append a block $(c_x,c_y,c_z,l_x,l_y,l_z,\kappa)$ to $L$;
    \WHILE{$U(0,1) \le p_{\text{nest}}$} 
        \STATE $l' \gets \max(l_x,l_y,l_z)/10$;
        \STATE $l_x\gets l_x+l'$; $l_y\gets l_y+l'$; $ l_z\gets l_z+l'$;
        \STATE Sample $\kappa \sim \mathsf{RandomDielectric}$;
        \STATE Append a block $(c_x,c_y,c_z,l_x,l_y,l_z,\kappa)$ to $L$;
    \ENDWHILE
\ENDFOR
\STATE Sample $\kappa \sim \mathsf{RandomDielectric}$;
\STATE Append a background block $(0, 0, 0, \infty,\infty,\infty, \kappa)$ to $L$;
\end{algorithmic}
\end{algorithm}

\subsection{Network Architecture for Transition Kernel Prediction}  \label{sec:model_architecture} 

The Poisson kernel is equivariant to the symmetry group of the cube, 
meaning that for any symmetry (including rotations, reflections, and inversions), the kernel values permute and transform consistently with the geometric transformation of the cube.

Thus, predicting this $\mathbf{p}_{\alpha}\in\mathbb{R}^{6\times N\times N}$ with a single model would introduce a high amount of redundancy across faces. A capacitance extraction typically requires millions of transitions, demanding a large number of neural network evaluations. This requirement makes model efficiency critical. Additionally, such a model would have to accurately predict both the shape of the Poisson kernel on each face as well as the overall probability across all faces. 

A two-stage prediction system avoids this redundancy and simplifies the learning complexity by first predicting a categorical distribution $\mathbf{F}\in \mathbb{R}^6$ where $\mathbf{F}_i=\sum_{j,k}(\mathbf{p}_{\alpha})_{i,j,k}$ across the six faces, and afterwards, predicting the conditional probabilities on the selected face $\frac{(\mathbf{p}_\alpha)_{i,:,:}}{\mathbf{F}_i} \in \mathbb{R}^{N\times N}$.

\begin{figure}[t]
    \centering
    \includegraphics[width=\linewidth]{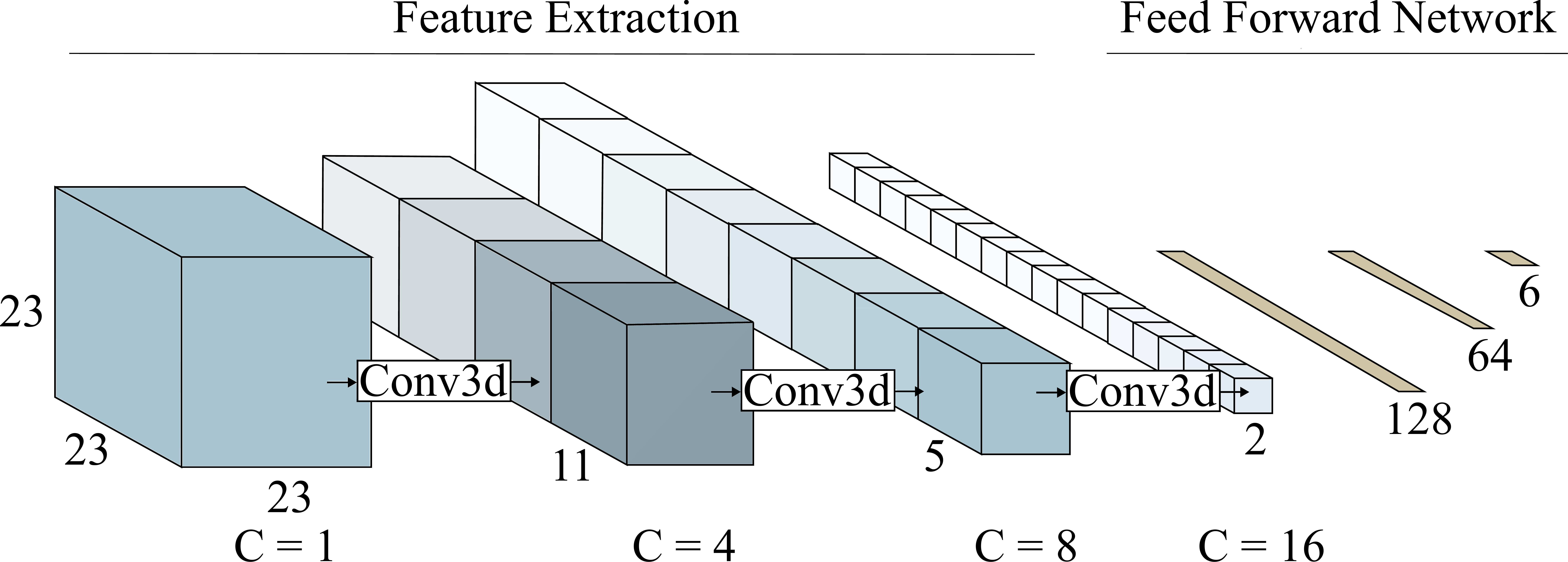}
    \caption{Face selector network architecture ($N=23$).}\label{fig:face_selector_architecture}
\end{figure}

The categorical distribution $\mathbf{F}$ requires modeling the interactions between the dielectrics within the transition cube. To capture these 3D spatial dependencies efficiently, we employ a 3D convolutional network referred to as the \textit{face selector} $\mathcal{F}_\theta$. It consists of four convolutional layers with a stride of 2 to progressively downsample the input dielectric tensor, as shown in Fig.~\ref{fig:face_selector_architecture}. The extracted features are then processed by a feed-forward network. The model output is normalized via softmax. The model is trained to minimize the KL divergence loss:
\begin{gather}
\mathcal{L}_{\text{face-select}}(\theta) = D_{KL}[\mathbf{F} \ \|\ \text{softmax}(\mathcal{F}_{\theta}(\mathcal{X}))].
\end{gather}

\begin{figure}
    \centering
    \includegraphics[width=1\linewidth]{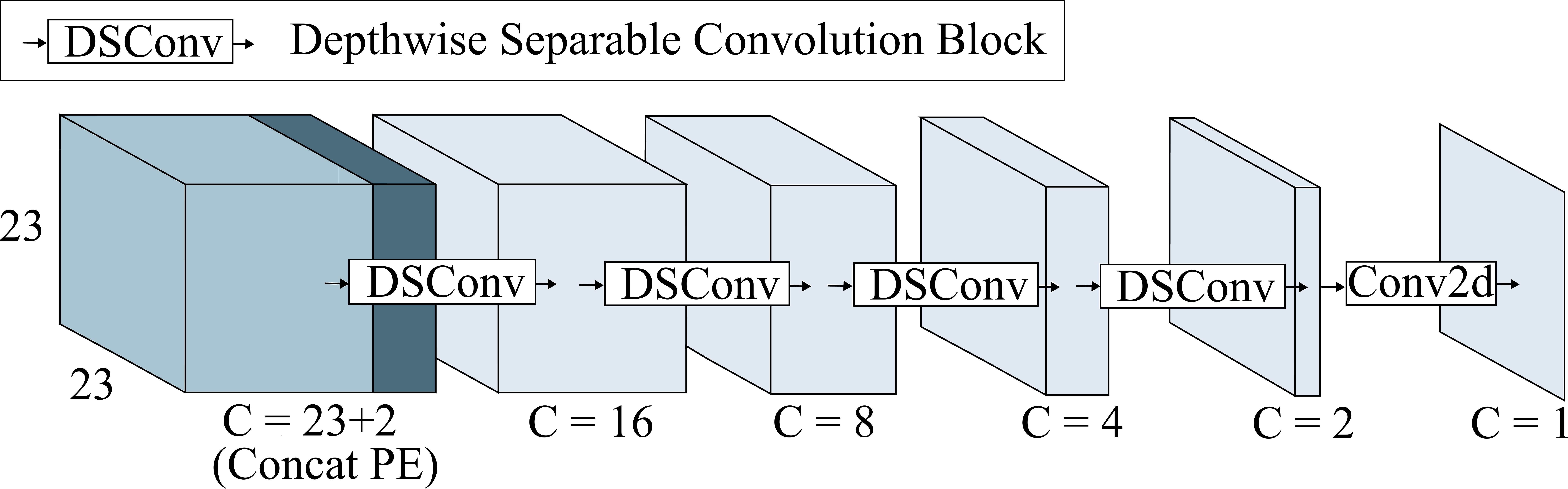}
    \caption{Face predictor network architecture ($N=23$).}
    \label{fig:face-predictor}
\end{figure}

For single-face Poisson kernel prediction, we treat the cube slices parallel to the target face as a stack of feature maps with $N$ channels.
The \textit{face predictor} $\mathcal{G}_{\theta}$ is implemented using 2D depthwise separable convolutions~\cite{mobilenet}, as shown in Fig. \ref{fig:face-predictor}.
The depthwise convolutions capture spatial patterns within each cube slice, while pointwise convolutions model inter-slice interactions across the channel dimension. This separation significantly reduces computational cost and aligns with the fact that the Poisson kernel's dependence on the dielectrics decays rapidly with distance from the surface. 

To provide explicit spatial context and assist the learning of the low-frequency components of the Poisson kernel, we concatenate the input with a simple grid-based positional encoding (PE), adding two extra channels $(x,y)$~\cite{liu2018intriguing}. This results in an input tensor $\tilde{\mathcal{X}} \in \mathbb{R}^{(N+2)\times N\times N}$.
Since Poisson kernel is inherently non-negative, the last layer of $\mathcal{G}_{\theta}$ is a ReLU activation.
Finally, the output of $\mathcal{G}_{\theta}$ is normalized using L1 normalization to ensure a valid probability distribution. The loss function is defined as:
\begin{gather}
    \mathcal{L}_{\text{face-predict}}(\theta)=D_{KL}\left[\frac{(\mathbf{p}_{\alpha})_{i,:,:}}{\mathbf{F}_i} \ \left\| \ \frac{\mathcal{G}_{\theta}(\tilde{\mathcal{X}})}{\left\|\mathcal{G}_{\theta}(\tilde{\mathcal{X}})\right\|_1}\right.\right]. 
\end{gather}

As discussed in Section~\ref{sec:frw}, the first transition in each walk requires a special set of quantities: the \textit{weight value} $w_\alpha$, the \textit{sign distribution} $\mathbf{s}_\alpha$ and the \textit{gradient kernel} $\mathbf{g}_\alpha$.
In practice, different components ($x$, $y$, or $z$) of the gradient can be requested, depending on the orientation of the Gaussian surface at the first sample point.

We only learn the $z$-component and derive the other components via cube symmetries, i.e., applying corresponding rotations and reflections to the dielectric input. Similarly, we employ a two-stage process, with a gradient face selector and gradient kernel predictors. 

The weight prediction constitutes a cumulative regression task similar to the face selector. We extend the output dimension of face selector $\mathcal{F}_{\theta}$ from 6 to 7 to simultaneously predict the face-wise distribution $\mathbf{F}^\nabla_i = \sum_{j,k}(\mathbf{g}_\alpha)_{i,j,k}$ and the weight magnitude $w_{\alpha} \in \mathbb{R}$. To accommodate this increased complexity, we scale up the network size appropriately (see Section \ref{sec:exp}). The training objective combines a KL divergence for face-wise probabilities with Mean Squared Error (MSE) for weight regression:
\begin{equation}
\begin{aligned}\label{eq:loss}
\mathcal{L}_{\text{grad-face-select}}(\theta) &=D_{KL}[\mathbf{F}^{\nabla} \ \| \ \text{softmax}(\mathcal{F}_{\theta}(\mathcal{X})_{1:6} )]\\
&+ \lambda |w_\alpha -\mathcal{F}_{\theta}(\mathcal{X})_7|^2.
\end{aligned}
\end{equation}

Given the structural similarity between the sign distribution $\mathbf{s}_\alpha$ and the gradient kernel $\mathbf{g}_\alpha$, we propose to learn the signed gradient kernel $\mathbf{s}_{\alpha}\mathbf{g}_{\alpha}$ rather than modeling them separately.
However, the gradient kernel is equivariant only to a subgroup of cube symmetries, necessitating at least two specialized face predictors to properly characterize the tangent (parallel to Gaussian surface) and normal (perpendicular to Gaussian surface) faces, as shown in Appendix A. The normal-face gradient kernels exhibit challenging learning characteristics including sign ambiguities and dual-peaked distributions, motivating the use of deeper network architectures (see Section \ref{sec:exp}). 
Since the signed target includes both positive and negative values, we do not employ the ReLU activation in the face predictor $\mathcal{G}_{\theta}$ and use MSE as the training loss:

\begin{gather}
\mathcal{L}_{\text{grad-face-predict}}(\theta) = \left| \frac{(\mathbf{s}_{\alpha}\mathbf{g}_{\alpha})_{i,:,:}}{\mathbf{F}^{\nabla}_i} - \frac{\mathcal{G}_{\theta}(\tilde{\mathcal{X}})}{\|\mathcal{G}_{\theta}(\tilde{\mathcal{X}})\|_1}\right|^2.
\end{gather}

\subsection{High-Throughput Inference Implementation}

Our implementation targets high-throughput capacitance extraction using (1) an  asynchronous producer-consumer architecture for continuous GPU utilization, (2) multi-instance model deployment for pipeline parallelism, and (3) custom CUDA kernels for efficient data processing.

\begin{figure}[!b]
    \centering
    \includegraphics[width=1\linewidth]{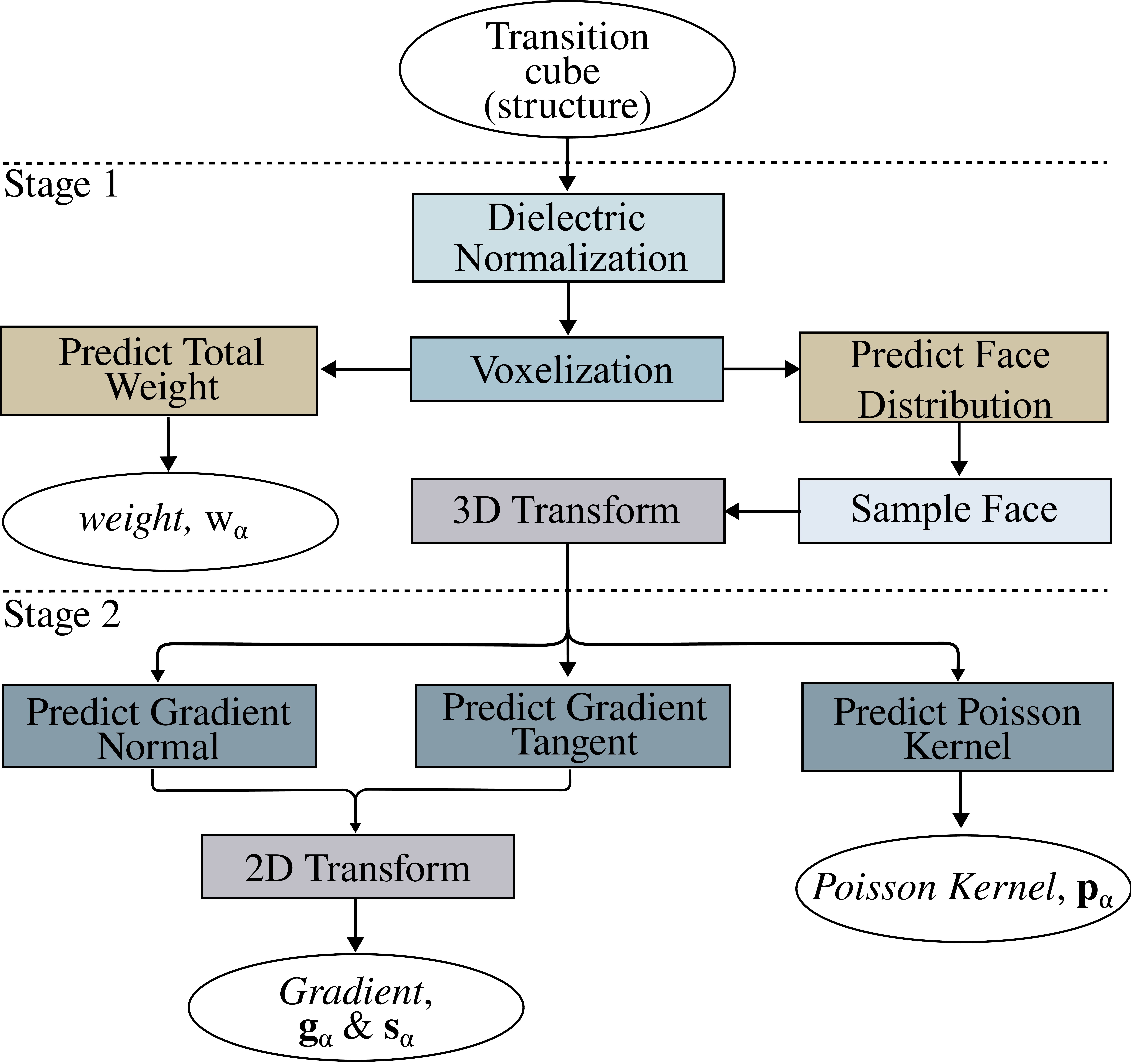}
    \caption{Transition quantities prediction pipeline.}
    \label{fig:gradient_sampling_flow}
\end{figure}

\paragraph{Batched GPU Processing Architecture}
When a walker encounters a non-stratified transition domain, the system employs the lightweight models described in Section~\ref{sec:model_architecture}, following the two-stage pipeline shown in Figure~\ref{fig:gradient_sampling_flow}. To maximize GPU utilization, transition sampling is batched. We implement a producer-consumer architecture where each thread manages a walker pool and submits transition tasks to a lock-free queue. Implementation details and pseudocode are provided in Appendix B.

\paragraph{Multi-Instance Deployment}
To reduce latency, we deploy multiple model instances enabling simultaneous GPU processing and CPU-GPU data transfers. We instantiate one Poisson solver for every two walker threads and one gradient solver shared across all walkers. Data is processed as soon as available, preventing walker threads from stalling while waiting for GPU results. 

\paragraph{Optimized Data Processing and Model Inference}
To maximize throughput, we minimize GPU memory transfers by transmitting compact structural descriptions rather than voxelized data, performing voxelization directly on GPU. Model inference uses TensorRT FP16 compilation and custom fused CUDA kernels for dielectric transformations.

\section{Experimental Results}

\subsection{Experimental Setup}

\begin{figure}[t]
    \centering
    \includegraphics[width=\linewidth]{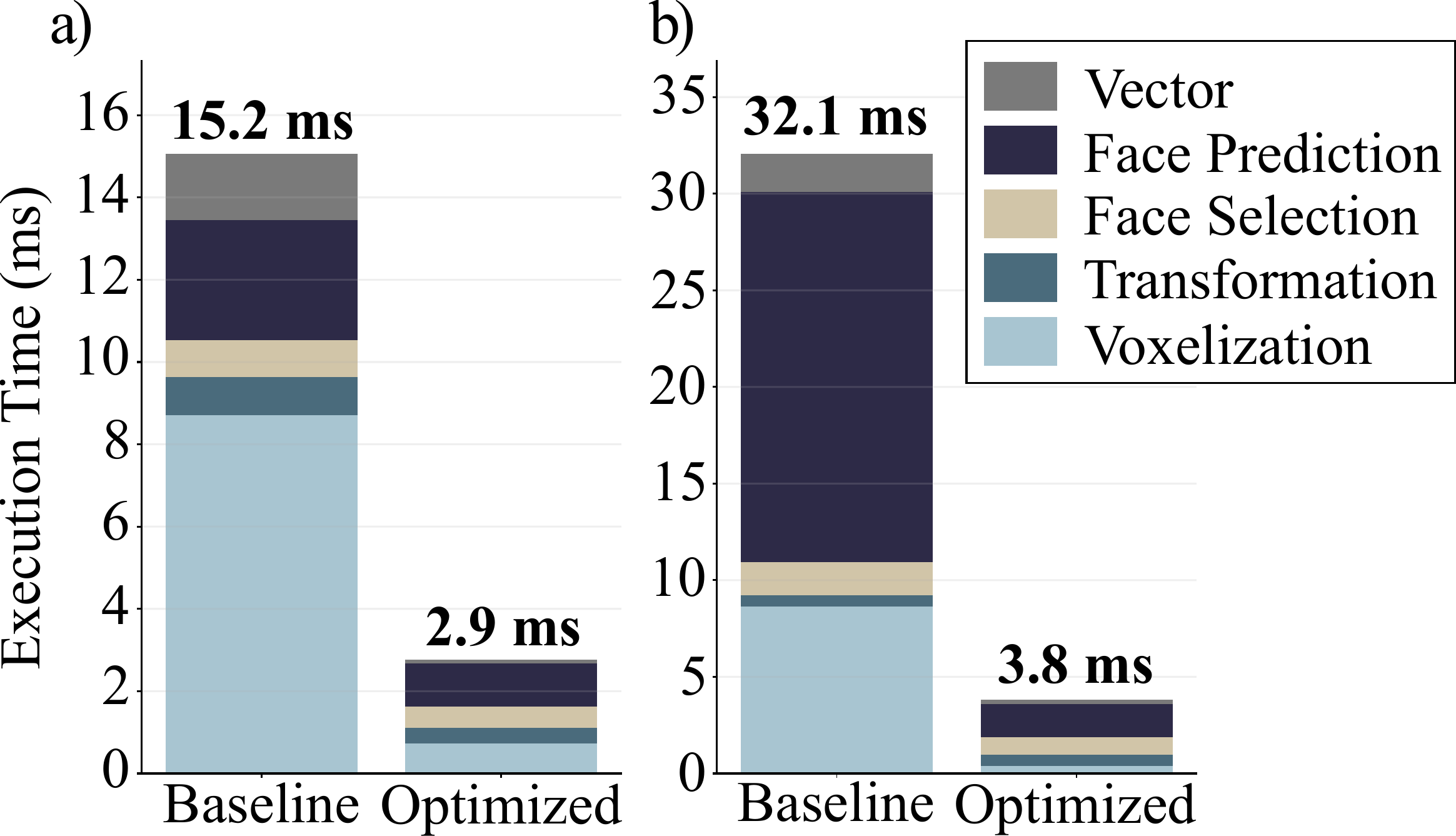}
    \caption{Performance analysis after introducing custom CUDA Kernels and TensorRT compilation, benchmarked with a batch size of 2048: (a) Poisson and (b) Gradient.}
    \label{fig:performance-comparison}
\end{figure}

All experiments were conducted on a server equipped with an Intel Xeon Silver 4214 CPU @ 2.20GHz and an NVIDIA RTX 4090 GPU. The transition cubes were discretized with $N = 23$. The TensorRT compilation and custom CUDA kernel optimizations described in Section 4.4 are critical for achieving practical deployment speeds as shown in Figure~\ref{fig:performance-comparison}.

\subsection{Model Training}\label{sec:exp}

\paragraph{Datasets.} We generated two comprehensive training datasets using the procedure described in Alg.~\ref{alg:dielectric} with $B=5$ and $p_{\text{nest}}=0.2$.
The first dataset contains 100,000 samples for Poisson kernel prediction. The second dataset comprises 100,000 samples for gradient kernel prediction. Both datasets were split 90-10 for training and validation. The FDM solver took 1.7 hours to generate the dataset.

\paragraph{Training configuration.} All models were trained using batch size of 16 for 200 epochs with the AdamW optimizer using $\beta_1 = 0.9$, $\beta_2 = 0.999$, and no weight decay. Gradient clipping was applied with a maximum norm of 1.0. The learning rate schedule employed cosine annealing from $1 \times 10^{-3}$ to $5 \times 10^{-6}$ with a warmup period during the first 20 epochs. For the dual-objective loss function in \eqref{eq:loss}, we used $\lambda = 1$. Each full training run required 12.3 hours.

\paragraph{Model Implementation Details.} The face solver architecture (see Fig. \ref{fig:face-predictor}) comprises a positional encoding layer followed by a channel projection using $1\times 1$ convolutions (from $23+2$ input channels to $16$ channels), four depthwise separable layers with channel progression $(16,16,8,4,2)$ and increasing dilation rates $(1, 1, 2, 3)$, and a final $1\times 1$ convolution head. The dilated convolutions enable the model to capture larger spatial contexts while maintaining computational efficiency and producing smoother output distributions. Each depthwise separable block consists of a $3\times 3$ depthwise convolution, batch normalization, GELU activation, followed by a $1 \times 1$ pointwise convolution, batch normalization, and GELU activation. The tangent gradient solver variant follows the same structure, and the normal gradient solver includes wider channels $(64,64,32,32,16,16,8,4)$. The 3D face selector networks (see Fig. \ref{fig:face_selector_architecture}) use standard 3D convolutions with a stride of 2, batch normalization, GELU activations, followed by fully connected layers. We use channel progressions of $(1,4,8,16)$ for Poisson kernels and $(1,8,16,64)$ for gradient kernels.

\begin{table}[b]
	\centering
	\begin{tabular}{l@{\hspace{2pt}}c@{\hspace{2pt}}c@{\hspace{2pt}}c@{\hspace{2pt}}c}
		\toprule
		\textbf{Model} & \textbf{Params} & \textbf{FLOPs} & \textbf{Loss} & \textbf{Valid. Loss}  \\
		\midrule
		Poisson Solver & 1.40 K & 0.84 M & KL & $2.1 \times 10^{-3}$ \\
		Gradient Tangent & 1.40 K & 0.84 M & MSE & $3.2 \times 10^{-8}$ \\
		Gradient Normal & 5.96 K & 3.48 M & MSE & $4.5 \times 10^{-8}$ \\
		Poisson Selector & 13.16 K & 0.31 M & KL & $7.6\times 10^{-4}$ \\
		Gradient Selector & 196.7 K & 1.16 M & KL+MSE & $2.8 \times 10^{-3}$ \\
		\bottomrule
	\end{tabular}
	\caption{Component Model Summary}
	\label{tab:training_results}
\end{table}

\begin{table*}[htbp]
	\centering
	\begin{threeparttable}
		{\setlength{\tabcolsep}{4pt}
		\begin{tabular}{ccrrrrrrrr}
			\toprule
			\multirow{2.5}{*}{\textbf{Case}} & 
			\multicolumn{1}{c}{\textbf{Node}} & 
			\multicolumn{1}{c}{\textbf{FRW-FDM}} & 
			\multicolumn{1}{c}{\textbf{GE-CNN}} & 
			\multicolumn{1}{c}{\textbf{FRW-AGF}} &
			\multicolumn{1}{c}{\textbf{Microwalk}} & 
			\multicolumn{4}{c}{\textbf{DeepRWCap (ours)}} \\ \cmidrule(lr){7-10}
			& 
			\multicolumn{1}{c}{{(nm)}} & 
			\multicolumn{1}{c}{\textbf{Error} (\%)} & 
			\multicolumn{1}{c}{\textbf{Error} (\%)} & 
			\multicolumn{1}{c}{\textbf{Error} (\%)} & 
			\multicolumn{1}{c}{\textbf{Error} (\%)} & 
			\multicolumn{1}{c}{\textbf{Error} (\%)} & 
			\multicolumn{1}{c}{\textbf{Tasks \tnote{a}}\, (M)} & 
			\multicolumn{1}{c}{\textbf{Poisson \tnote{b}}\, (\%)} & 
			\multicolumn{1}{c}{\textbf{Grad. \tnote{b}}\, (\%)} \\
			\midrule
			1  & 16 & $0.4 \pm 0.2$ & $4.9 \pm 0.2$ & $0.8 \pm 0.1$ & $0.6 \pm 0.2$ & $1.2 \pm 0.1$ & 23.20 & 54.7 & 11.8 \\
			2  & 16 & $1.1 \pm 0.3$ & \underline{$7.0 \pm 0.4$} & $1.8 \pm 0.3$ & $1.7 \pm 0.3$ & $2.1 \pm 0.3$ & 5.79  & 27.4 & 4.4  \\
			3  & 16 & $1.0 \pm 0.1$ & \underline{$5.6 \pm 0.2$} & $2.2 \pm 0.1$ & $1.8 \pm 0.2$ & $1.2 \pm 0.1$ & 7.08  & 55.8 & 13.4 \\
			4  & 28 & $1.8 \pm 0.4$ & \underline{$6.6 \pm 0.3$} & $1.4 \pm 0.4$ & $0.5 \pm 0.4$ & $0.7 \pm 0.3$ & 41.07 & 27.7 & 2.6  \\
			5  & 55 & $1.0 \pm 0.3$ & $3.9 \pm 0.1$ & $0.7 \pm 0.2$ & $0.9 \pm 0.3$ & $2.2 \pm 0.2$ & 57.64 & 15.5 & 0.0  \\
			6  & 55 & $0.4 \pm 0.2$ & \underline{$7.0 \pm 0.2$} & $2.0 \pm 0.3$ & $0.4 \pm 0.3$ & $0.9 \pm 0.3$ & 41.67 & 26.3 & 0.0  \\
			7  & 16 & $2.2 \pm 0.7$ & $1.0 \pm 0.3$ & $1.7 \pm 0.5$ & $2.0 \pm 0.6$ & $1.2 \pm 0.4$ & 4.18  & 11.2 & 1.3  \\
			8  & 16 & $0.7 \pm 0.6$ & $0.9 \pm 0.3$ & $0.3 \pm 0.3$ & $0.6 \pm 0.5$ & $0.6 \pm 0.5$ & 2.33  & 10.9 & 1.5  \\
			9  & 12 & $1.5 \pm 0.8$ & \underline{$22.9 \pm 0.7$} & \underline{$17.0 \pm 1.0$} & $0.9 \pm 0.5$ & $1.1 \pm 0.9$ & 12.02 & 13.8 & 2.3  \\
			10 & 12 & $0.6 \pm 0.6$ & \underline{$27.1 \pm 0.8$} & \underline{$23.9 \pm 1.3$} & $0.6 \pm 0.4$ & $1.2 \pm 0.9$ & 6.23  & 14.3 & 3.1  \\
			\bottomrule
		\end{tabular}}
		\caption{Performance Comparison of Random Walk Methods (with relative errors larger than 5\% underlined).}
		\label{tab:merged_comparison}
		\begin{tablenotes}
			\footnotesize
			\item[a] Average number of transition cubes solved by DeepRWCap.
			\item[b] Average percentage of transition cubes classified as non-stratified, requiring neural inference for kernel sampling.
		\end{tablenotes}
	\end{threeparttable}
\end{table*}

\begin{figure*}[ht]
    \centering
    \includegraphics[width=\textwidth]{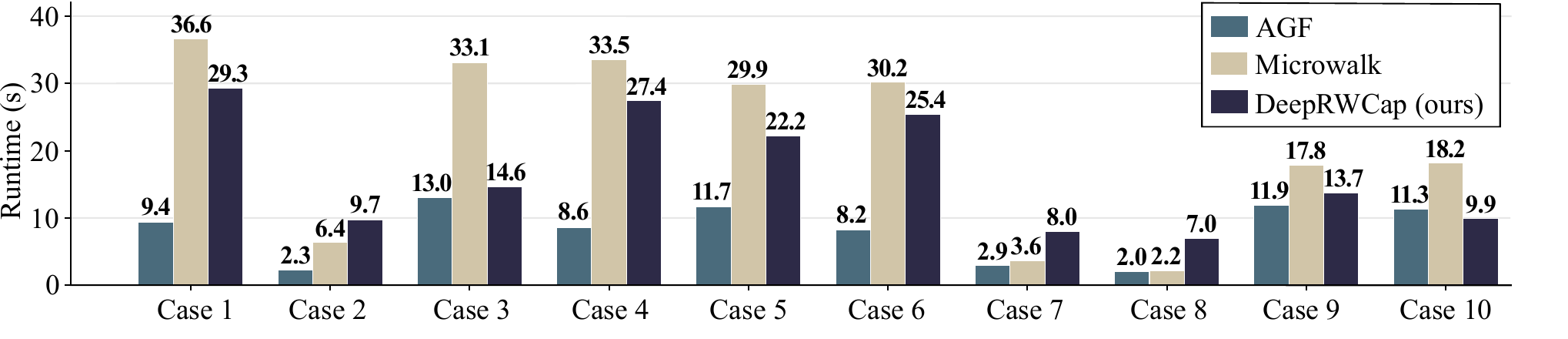}
    \caption{Method Runtime Comparison}
    \label{fig:runtime_comparison}
\end{figure*}

\subsection{Capacitance Extraction Results}

Cases 1-6 involve simple parallel-wire structures across different metal layers and technologies. Cases 7-10 are FinFET structures with local interconnects, with cases 9-10 being particularly challenging due to high-contrast conformal dielectrics. For all cases, the FRW process terminates when the estimated stochastic error falls below 1\%. Ground-truth values are obtained from Raphael, which takes hours per case. A detailed breakdown of test case structures and example visualizations is provided in Appendix C.

DeepRWCap was configured with 8 walker threads (512 walkers per thread), 4 threads for Poisson sampling, and 1 thread for gradient sampling. The CPU-based FRW-FDM, FRW-AGF~\cite{huang2024enhancing} and Microwalk~\cite{huang2025efficientfrwtransitionsstochastic} used 16 walker threads. For a fair comparison, the GE-CNN approach~\cite{visvardis2023deep} was adapted to the same GPU-accelerated architecture as DeepRWCap. 

Table~\ref{tab:merged_comparison} compares method accuracy across test cases, with each method evaluated over 10 independent trials. The percentage of non-stratified domains requiring neural network inference varies significantly: Poisson kernel sampling in 11.2-55.8\% of domains and gradient kernel sampling in 0.0-13.4\% of domains. 
Runtime comparisons are shown in Figure~\ref{fig:runtime_comparison}. Per-case runtimes for all methods, including FRW-FDM and GE-CNN, are listed in Appendix D. Overall, cases 7-10 present shorter runtimes because the smaller conductor spacing accelerates FRW convergence.
Statistical analysis using Wilcoxon signed-rank tests ($\alpha = 0.05$) shows DeepRWCap achieves a $1.23\times$ speedup over Microwalk ($p = 0.024$) while maintaining statistically equivalent accuracy. Although AGF achieved faster execution times, it exhibited poor reliability with high variance in error rates ($5.18\% \pm 7.81\%$ vs DeepRWCap's $1.24\% \pm 0.53\%$).

\subsection{Ablation Study}

We conduct an ablation study to assess the impact of architectural design choices on accuracy and efficiency for single-face Poisson kernel prediction. We evaluate model variants under consistent training settings in Section \ref{sec:exp}. The MLP baseline includes three hidden layers of 2048 units each. We also evaluate with standard 2D convolutions. As a third baseline, we replace the depthwise separable convolution layers with 3D convolutional layers with stride-2 downsampling to reduce depth while preserving width and height. Architectural specifications for all ablation models are provided in Appendix E.

\begin{table}[t]
	\centering
	{\setlength{\tabcolsep}{1mm}
	\begin{tabular}{lrrrr}
\toprule
\textbf{Architecture} & \textbf{Params} & \textbf{FLOPs} & \textbf{L2} (\%) & \textbf{KL Div} \\
\midrule
MLP & 34.4 M & 34.4 M & 7.83 & 0.0133 \\
3D Conv & 19.7 K & 20.3 M & 24.15 & 0.0298 \\
2D Conv & 4.28 K & 2.31 M & 13.80 & 0.0125 \\
GE-CNN + GMM & 0.43 M & 0.65 M & 26.63 & 0.0403\\
DS Conv & 1.37 K & 0.82 M & 12.15 & 0.0083 \\
DS Conv + Learn PE & 2.46 K & 0.84 M & 4.11 & 0.0023\\
DS Conv + Grid PE & 1.40 K & 0.84 M & 3.93 & 0.0021\\
\bottomrule
\end{tabular}}
\caption{Model Architecture Ablation}
\label{tab:model_ablation}
\end{table}

From Table~\ref{tab:model_ablation}, depthwise separable convolutional models achieve superior accuracy with significantly fewer parameters and FLOPs compared to MLPs, 2D convolutions, and 3D convolutions. The 3D convolutional approach, while conceptually appealing for processing volumetric dielectric data, proves computationally inefficient with substantially higher parameter count and FLOPs. This validates our design choice of decomposing the 3D problem into face selection and 2D face-specific kernel prediction. Moreover, incorporating positional encodings notably improves performance, with the fixed grid encoding yielding the best accuracy and lowest KL divergence.

\section{Conclusion}

We presented DeepRWCap, a neural-guided random walk solver that accelerates capacitance extraction for advanced IC designs. Our approach combines a two-stage CNN architecture with GPU-optimized inference to predict transition kernels in multi-dielectric domains. The compact depthwise separable networks with positional encoding achieve high accuracy while maintaining computational efficiency. Evaluated on 10 industrial test cases spanning 12-55 nm technologies, DeepRWCap demonstrates a mean relative error of 1.24\% with significant speedups over state-of-the-art methods—achieving up to 49\% acceleration on complex designs. The framework provides a practical solution for capacitance extraction in modern semiconductor technologies and serves as a basis for future extensions to other elliptic PDEs.

\section*{Acknowledgments}
This work was partially supported by National Science and Technology Major Project (2021ZD0114703), Beijing Natural Science Foundation (Z220003), and the MIND project (MINDXZ202406). W. Yu is the corresponding author.
\bibliography{references}

@article{le1992stochastic,
	title={{A stochastic algorithm for high speed capacitance extraction in integrated circuits}},
	author={{Y. L. {Le Coz} and R. B. Iverson}},
	journal={Solid-State Electron.},
	volume={35},
	number={7},
	pages={1005--1012},
	year={1992},
	publisher={Elsevier}
}

@article{yu2013rwcap,
	title={{RWCap: A floating random walk solver for 3-D capacitance extraction of very-large-scale integration interconnects}},
	author={Yu, W. and Zhuang, H. and Zhang, C. and Hu, G. and Liu, Z.},
	journal={IEEE Trans. Comput.-Aided Des. Integr. Circuits Syst.},
	volume={32},
	number={3},
	pages={353--366},
	year={2013},
	publisher={IEEE}
}

@article{yang2020floating,
	title={Floating random walk capacitance solver tackling conformal dielectric with on-the-fly sampling on eight-octant transition cubes},
	author={Yang, M. and Yu, W.},
	journal={IEEE Trans. Comput.-Aided Des. Integr. Circuits Syst.},
	volume={39},
	number={12},
	pages={4935--4943},
	year={2020},
	publisher={IEEE}
}

@ARTICLE{visvardis2023deep,
  author={Visvardis, M. and Liaskovitis, P. and Efstathiou, E.},
  journal={IEEE Trans. Comput.-Aided Des. Integr. Circuits Syst.}, 
  title={Deep-learning-driven random walk method for capacitance extraction}, 
  year={2023},
  volume={42},
  number={8},
  pages={2643-2656}
}

@inproceedings{yu2021advancements,
  title={Advancements and challenges on parasitic extraction for advanced process technologies},
  author={Yu, W. and Song, M. and Yang, M.},
  booktitle={Proc. ASP-DAC},
  pages={841--846},
  year={2021}
}

@Article{Ivica2021,
 author={Ivica Smolić and Bruno Klajn},
 title={CAPACITANCE MATRIX REVISITED},
 volume={92},
 journal={Progress In Electromagnetics Research B},
 year={2021},
 pages={1-18},
 doi={10.2528/PIERB21011501}
}

@article{huang2024floating,
  title={The Floating Random Walk Method With Symmetric Multiple-Shooting Walks for Capacitance Extraction},
  author={Huang, Jiechen and Yang, Ming and Yu, Wenjian},
  journal={IEEE Trans. Comput.-Aided Des. Integr. Circuits Syst.},
  year={2024},
  volume={43},
  number={7},
  pages={2098-2111}
}

@inproceedings{huang2024enhancing,
  title={Enhancing {3-D} Random Walk Capacitance Solver with Analytic Surface {Green's} Functions of Transition Cubes},
  author={Huang, Jiechen and Yu, Wenjian},
  booktitle={Proc. DAC},
  year={2024}
}

@article{kakutani1944143,
  title={Two-dimensional brownian motion and harmonic functions},
  author={Kakutani, Shizuo},
  journal={Proceedings of the Imperial Academy},
  volume={20},
  number={10},
  pages={706--714},
  year={1944}
}

@article{muller1956some,
  title={Some continuous {Monte Carlo} methods for the {Dirichlet} problem},
  author={Muller, Mervin E},
  journal={The Annals of Mathematical Statistics},
  pages={569--589},
  year={1956},
  publisher={JSTOR}
}

@book{oksendal2010stochastic,
  title={Stochastic Differential Equations: An Introduction with Applications, Sixth Edition},
  author={{\O}ksendal, B.},
  year={2003},
  publisher={Springer}
}

@article{huang2025efficientfrwtransitionsstochastic,
      title={Efficient {FRW} Transitions via Stochastic Finite Differences for Handling Non-Stratified Dielectrics}, 
      author={Jiechen Huang and Wenjian Yu},
  journal={IEEE Trans. Comput.-Aided Des. Integr. Circuits Syst.},
  volume={},
  number={},
  pages={},
  year={2025},
  publisher={IEEE}
}

@inproceedings{nam2024poisson,
author = {Nam, Hong Chul and Berner, Julius and Anandkumar, Anima},
title = {Solving poisson equations using neural walk-on-spheres},
year = {2024},
booktitle = {Proceedings of the 41st International Conference on Machine Learning},
articleno = {1513},
numpages = {16},
location = {Vienna, Austria},
series = {ICML'24}
}

@article{miller2024differential,
author = {Miller, Bailey and Sawhney, Rohan and Crane, Keenan and Gkioulekas, Ioannis},
title = {Differential Walk on Spheres},
year = {2024},
issue_date = {December 2024},
publisher = {Association for Computing Machinery},
address = {New York, NY, USA},
volume = {43},
number = {6},
issn = {0730-0301},
url = {https://doi.org/10.1145/3687913},
doi = {10.1145/3687913},
journal = {ACM Trans. Graph.},
month = nov,
articleno = {174},
numpages = {18}
}

@article{sawhney2020monte,
  title={Monte Carlo geometry processing: A grid-free approach to PDE-based methods on volumetric domains},
  author={Sawhney, Rohan and Crane, Keenan},
  journal={ACM Trans. Graph.},
  volume={39},
  number={4},
  year={2020}
}

@inproceedings{li2023neural,
  title={Neural caches for monte carlo partial differential equation solvers},
  author={Li, Zilu and Yang, Guandao and Deng, Xi and De Sa, Christopher and Hariharan, Bharath and Marschner, Steve},
  booktitle={SIGGRAPH Asia 2023 Conference Papers},
  pages={1--10},
  year={2023}
}

@book{Griffiths_2023, place={Cambridge}, edition={5}, title={Introduction to Electrodynamics}, publisher={Cambridge University Press}, author={Griffiths, David J.}, year={2023}}

@book{axler2001harmonic,
  title={Harmonic Function Theory},
  author={Axler, S. and Bourdon, P. and Wade, R.},
  isbn={9780387952185},
  lccn={00053771},
  series={Graduate Texts in Mathematics},
  url={https://books.google.com/books?id=wATLzBfup-wC},
  year={2001},
  publisher={Springer}
}

@article{sawhney2022grid,
author = {Sawhney, Rohan and Seyb, Dario and Jarosz, Wojciech and Crane, Keenan},
title = {Grid-free Monte Carlo for PDEs with spatially varying coefficients},
year = {2022},
issue_date = {July 2022},
publisher = {Association for Computing Machinery},
address = {New York, NY, USA},
volume = {41},
number = {4},
issn = {0730-0301},
url = {https://doi.org/10.1145/3528223.3530134},
doi = {10.1145/3528223.3530134},
journal = {ACM Trans. Graph.},
month = jul,
articleno = {53},
numpages = {17}
}

@article{raissi2019physics,
  title={Physics-informed neural networks: A deep learning framework for solving forward and inverse problems involving nonlinear partial differential equations},
  author={Raissi, Maziar and Perdikaris, Paris and Karniadakis, George E},
  journal={Journal of Computational Physics},
  volume={378},
  pages={686--707},
  year={2019},
  publisher={Elsevier}
}

@article{mobilenet,
  author       = {Andrew G. Howard and
                  Menglong Zhu and
                  Bo Chen and
                  Dmitry Kalenichenko and
                  Weijun Wang and
                  Tobias Weyand and
                  Marco Andreetto and
                  Hartwig Adam},
  title        = {MobileNets: Efficient Convolutional Neural Networks for Mobile Vision
                  Applications},
  journal      = {CoRR},
  volume       = {abs/1704.04861},
  year         = {2017},
  url          = {http://arxiv.org/abs/1704.04861},
  eprinttype    = {arXiv},
  eprint       = {1704.04861},
  bibsource    = {dblp computer science bibliography, https://dblp.org}
}

@article{yang2022cnncap,
  title={CNN-Cap: Effective Convolutional Neural Network Based Capacitance Models for Interconnect Capacitance Extraction},
  author={Yang, Dingcheng and Li, Haoyuan and Yu, Wenjian and Guo, Yuanbo and Liang, Wenjie},
  journal={ACM Transactions on Design Automation of Electronic Systems},
  volume={28},
  number={4},
  pages={1--22},
  year={2023},
  publisher={ACM},
  doi={10.1145/3564931}
}

@inproceedings{cai2024pctcap,
  title={PCT-Cap: Point Cloud Transformer for Accurate 3D Capacitance Extraction},
  author={Cai, Ye and Liang, Yuyao and Luo, Zhipeng and Xie, Biwei and Li, Xingquan},
  booktitle={Proceedings of the 2024 2nd International Symposium of Electronics Design Automation (ISEDA)},
  pages={421--426},
  year={2024},
  publisher={IEEE},
  doi={10.1145/3658617.3703148}
}

@article{liu2024gnncap,
  title={GNN-Cap: Chip-Scale Interconnect Capacitance Extraction Using Graph Neural Network},
  author={Liu, Lin and Yang, Fei and Shang, Li and Zeng, Xiangyu},
  journal={IEEE Transactions on Computer-Aided Design of Integrated Circuits and Systems},
  volume={43},
  number={4},
  pages={1206--1217},
  year={2024},
  publisher={IEEE}
}

@inproceedings{gomes2022ponte,
  title={Ponte Vecchio: A Multi-Tile 3D Stacked Processor for Exascale Computing},
  author={Gomes, W. and Koker, A. and Stover, P. and Ingerly, D. and Siers, S. and Venkataraman, S. and Pelto, C. and Shah, T. and Rao, A. and O'Mahony, F. and Karl, E. and Cheney, L. and Rajwani, I. and Jain, H. and Cortez, R. and Chandrasekhar, A. and Kanthi, B. and Koduri, R.},
  booktitle={IEEE International Solid-State Circuits Conference (ISSCC)},
  pages={42--44},
  year={2022},
  organization={IEEE}
}

@inproceedings{wuu2022vcache,
  title={3D V-Cache: The Implementation of a Hybrid-Bonded 64MB Stacked Cache for a 7nm x86-64 CPU},
  author={Wuu, J. and Agarwal, R. and Ciraula, M. and Dietz, C. and Johnson, B. and Johnson, D. and Schreiber, R. and Swaminathan, R. and Walker, W. and Naffziger, S.},
  booktitle={IEEE International Solid-State Circuits Conference (ISSCC)},
  pages={428--429},
  volume={2022},
  year={2022},
  organization={IEEE}
}

@article{mukesh2022review,
  title={A review of the gate-all-around nanosheet FET process opportunities},
  author={Mukesh, Sagarika and Zhang, Jingyun},
  journal={Electronics},
  volume={11},
  number={21},
  pages={3589},
  year={2022},
  publisher={MDPI}
}

@inproceedings{liebmann2021cfet,
  title={CFET design options, challenges, and opportunities for 3D integration},
  author={Liebmann, L and Smith, J and Chanemougame, D and Gutwin, P},
  booktitle={2021 IEEE International Electron Devices Meeting (IEDM)},
  pages={3--1},
  year={2021},
  organization={IEEE}
}

@inproceedings{zhong2024preroutgnn,
  title={Preroutgnn for timing prediction with order preserving partition: Global circuit pre-training, local delay learning and attentional cell modeling},
  author={Zhong, Ruizhe and Ye, Junjie and Tang, Zhentao and Kai, Shixiong and Yuan, Mingxuan and Hao, Jianye and Yan, Junchi},
  booktitle={Proceedings of the AAAI Conference on Artificial Intelligence},
  volume={38},
  pages={17087--17095},
  year={2024}
}

@article{cheng2022policy,
  title={The policy-gradient placement and generative routing neural networks for chip design},
  author={Cheng, Ruoyu and Lyu, Xianglong and Li, Yang and Ye, Junjie and Hao, Jianye and Yan, Junchi},
  journal={Advances in Neural Information Processing Systems},
  volume={35},
  pages={26350--26362},
  year={2022}
}

@inproceedings{lai2025analogcoder,
  title={Analogcoder: Analog circuit design via training-free code generation},
  author={Lai, Yao and Lee, Sungyoung and Chen, Guojin and Poddar, Souradip and Hu, Mengkang and Pan, David Z and Luo, Ping},
  booktitle={Proceedings of the AAAI Conference on Artificial Intelligence},
  volume={39},
  pages={379--387},
  year={2025}
}

@inproceedings{worrall2018cubenet,
  title={Cubenet: Equivariance to 3d rotation and translation},
  author={Worrall, Daniel and Brostow, Gabriel},
  booktitle={Proceedings of the European Conference on Computer Vision (ECCV)},
  pages={567--584},
  year={2018}
}

@article{liu2018intriguing,
  title={An intriguing failing of convolutional neural networks and the coordconv solution},
  author={Liu, Rosanne and Lehman, Joel and Molino, Piero and Petroski Such, Felipe and Frank, Eric and Sergeev, Alex and Yosinski, Jason},
  journal={Advances in neural information processing systems},
  volume={31},
  year={2018}
}

\clearpage

\setcounter{figure}{0}
\setcounter{table}{0}
\setcounter{algorithm}{0}

\appendix

\section{Additional Details for Gradient Kernel Prediction}

\begin{figure}[h]
    \centering
    \includegraphics[width=\linewidth]{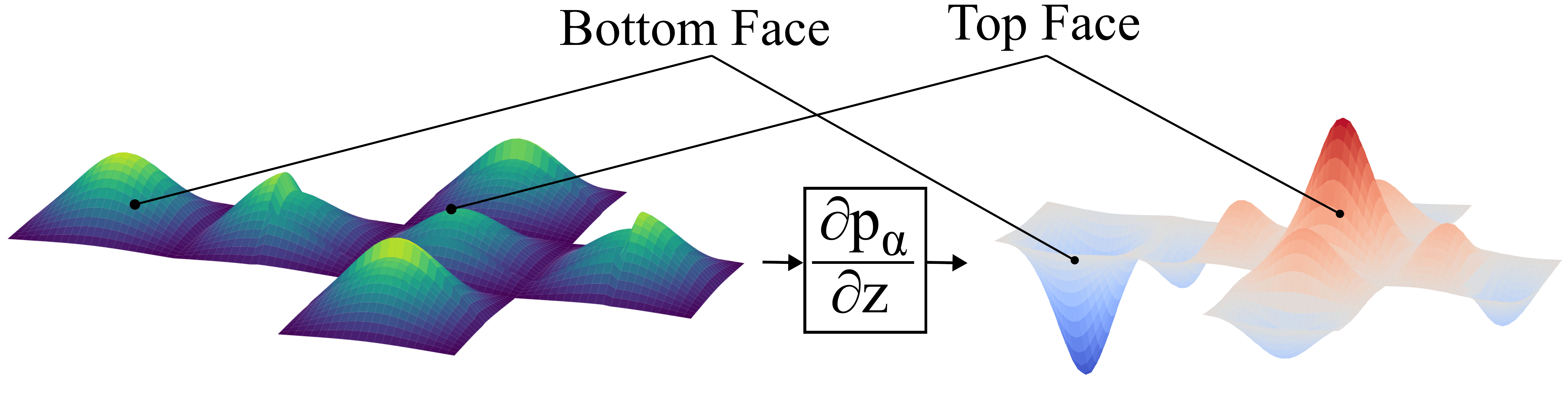}
    \caption{Unfolded Poisson kernel (left) and gradient kernel (right).}
    \label{fig:gradient_tangent_normal}
\end{figure}

The face Poisson kernel maintains the same fundamental shape regardless of face orientation, but the gradient kernel exhibits significantly different characteristics depending on the face's position and orientation, with its structure varying dramatically based on whether we consider faces that are tangential or normal to the gradient direction, as illustrated in Figure~\ref{fig:gradient_tangent_normal}. 

For faces oriented \textit{tangentially} to the gradient direction (such as the top and bottom faces when computing the gradient with respect to the $z$-axis), the gradient kernels are symmetric, differing only in sign. However, for faces oriented \textit{normal} to the gradient direction (the side faces), the gradient kernel exhibits a fundamentally different structure characterized by two distinct peaks: a positive peak at one end and a negative peak at the other. This bipolar structure makes the prediction task significantly more challenging, as the model must accurately capture both the magnitude and spatial distribution of these opposing peaks while preserving the correct sign relationships.

The increased complexity of gradient kernel prediction for normal-oriented faces is quantified in Table~\ref{tab:gradient_loss_comparison}. We trained our smallest model configuration using L1 output normalization with the absolute values while preserving the sign on three different tasks: Poisson kernel prediction on face 0, gradient kernel prediction on face 1 (tangential orientation), and gradient kernel prediction on face 2 (normal orientation). The results demonstrate that the prediction loss on the validation set for the normal-oriented face is approximately one order of magnitude higher than for the tangential face and the Poisson kernel task, confirming the fundamental difficulty in modeling the bipolar gradient structure. This substantial increase in prediction difficulty for normal-oriented faces necessitated the development of a larger model architecture.

\begin{table}[hb]
\centering
\begin{threeparttable}
\begin{tabular}{@{}lcc@{}}
\toprule
\textbf{Kernel}  & \textbf{MSE Loss} & \textbf{Rel. L2 Loss} (\%) \\
\midrule
Poisson                       & $4.20\times 10^{-8}$ & $5.64$ \\
Gradient (\textit{Tangential})& $4.67\times 10^{-8}$ & $5.34$ \\
Gradient (\textit{Normal})    & $1.94\times 10^{-7}$ & $11.83$ \\
\bottomrule
\end{tabular}
\end{threeparttable}
\caption{Gradient kernel prediction loss comparison across different tasks using the smallest model with L1 output normalization (trained with batch size of 256).}
\label{tab:gradient_loss_comparison}
\end{table}

\section{DeepRWCap Implementation Details}

\begin{algorithm}[htbp]
\caption{Producer: Random Walker Thread}
\label{alg:parallel_random_walk}
\textbf{Input}: Walker count $N_w$, termination criterion\\
\textbf{Output}: Capacitance matrix $\hat{\mathbf{C}}$
\begin{algorithmic}[1]
\STATE Initialize a walkers pool $P$ of $N_w$ walkers
\STATE Initialize samplers $Q_{\text{gradient}}$ and $Q_{\text{poisson}}$
\WHILE{termination criterion is not met}
    \FOR{each walker $p \in P$}
        \IF{$p.\mathsf{state} = \mathsf{Terminated}$}
            \STATE Update $\hat{\mathbf{C}}$ with the weight $\{w_\alpha,\mathbf{s}_\alpha\}$
            \STATE $p.\mathsf{state} \gets \mathsf{Active}$
        \ELSIF{$p.\mathsf{state} = \mathsf{WaitingForGPU}$}
            \IF{$\mathsf{IsDataAvailable}(p)$}
                \IF{$\mathsf{IsFirstStep}(p)$}
                    \STATE $\{\mathbf{v}_\alpha, w_\alpha\} \gets \mathsf{RetrieveResult}(p,Q_{\text{gradient}})$ 
                    \STATE $\mathsf{PerformTransition}(p, \mathbf{v}_\alpha, w_\alpha)$
                \ELSE
                    \STATE $\mathbf{v}_\alpha \gets \mathsf{RetrieveResult}(p,Q_{\text{poisson}})$
                    \STATE $\mathsf{PerformTransition}(p, \mathbf{v}_\alpha)$
                \ENDIF 
                \IF{$p$ hits a conductor}
                    \STATE $p.\mathsf{state} \gets \mathsf{Terminated}$
                \ELSE
                    \STATE $p.\mathsf{state} \gets \mathsf{Active}$
                \ENDIF
            \ENDIF
        \ELSIF{$p.\mathsf{state} = \mathsf{Active}$}
            \STATE $S \gets \mathsf{GetCurrentTransitionCube}(p)$
            \IF{$\mathsf{IsStratified}(S)$}
                \STATE $\mathsf{PerformTransitionAGF}(p, S)$
                \IF{$p$ hits a conductor}
                    \STATE $p.\mathsf{state} \gets \mathsf{Terminated}$
                \ENDIF
            \ELSE
                \IF{$\mathsf{IsFirstStep}(p)$}
                    \STATE $\mathsf{Submit}(\{p, S\}, Q_{\text{gradient}})$
                \ELSE
                    \STATE $\mathsf{Submit}(\{p, S\}, Q_{\text{poisson}})$
                \ENDIF
                \STATE $p.\mathsf{state} \gets \mathsf{WaitingForGPU}$
            \ENDIF
        \ENDIF
    \ENDFOR
\ENDWHILE
\end{algorithmic}
\end{algorithm}

The DeepRWCap inference process operates through a producer-consumer architecture where random walker threads (producers) generate sampling tasks and sampler threads (consumers) provide transition samples to guide the next step of the walk. The sampler threads handle GPU data transfers, synchronization, and CUDA kernel launches for batched inference. Algorithm~\ref{alg:parallel_random_walk} details the producer thread execution, with the following key function definitions:

\textbf{Walker State Management:}
\begin{itemize}
\item $\mathsf{IsDataAvailable}(p)$: Checks if sampling results are ready for walker $p$.
\item $\mathsf{IsFirstStep}(p)$: Determines if walker $p$ is at its initial position (requiring gradient kernel).
\item $\mathsf{GetCurrentTransitionCube}(p)$: Extracts the local dielectric configuration around walker $p$.
\item $\mathsf{IsStratified}(S)$: Tests if dielectric cube $S$ contains only stratified materials.
\end{itemize}

\textbf{Task Submission and Retrieval:}
\begin{itemize}
\item $\mathsf{Submit}(\{p, S\}, Q)$: Submits walker $p$ and dielectric configuration $S$ to the appropriate sampler queue $Q$.
\item $\mathsf{RetrieveResult}(p, Q)$: Retrieves processed results from sampler queue $Q$ for walker $p$. Returns $\{\mathbf{v}_\alpha, w_\alpha\}$ (transition vector and weight) for gradient sampling or $\mathbf{v}_\alpha$ (transition vector only) for Poisson sampling.
\end{itemize}

\textbf{Transition Execution:}
\begin{itemize}
\item $\mathsf{PerformTransition}(p, \mathbf{v}_\alpha, w_\alpha)$: Updates walker $p$'s position using transition vector $\mathbf{v}_\alpha$ and records the weight $w_\alpha$ for first steps.
\item $\mathsf{PerformTransition}(p, \mathbf{v}_\alpha)$: Updates walker $p$'s position using transition vector $\mathbf{v}_\alpha$ for subsequent steps.
\item $\mathsf{PerformTransitionAGF}(p, S)$: Executes analytical transition using the AGF approach \cite{huang2024enhancing} for stratified regions.
\end{itemize}

\textbf{Queue Management:}
The algorithm maintains separate queues for the gradient sampler ($Q_{\text{gradient}}$) and the Poisson sampler ($Q_{\text{poisson}}$), enabling specialized batch processing for each kernel type. Walkers transition between $\mathsf{Active}$, $\mathsf{WaitingForGPU}$, and $\mathsf{Terminated}$ states as they progress through the dielectric structure until reaching conductor boundaries.

\section{Test Case Description}

\begin{figure}
    \centering
    \includegraphics[width=1\linewidth]{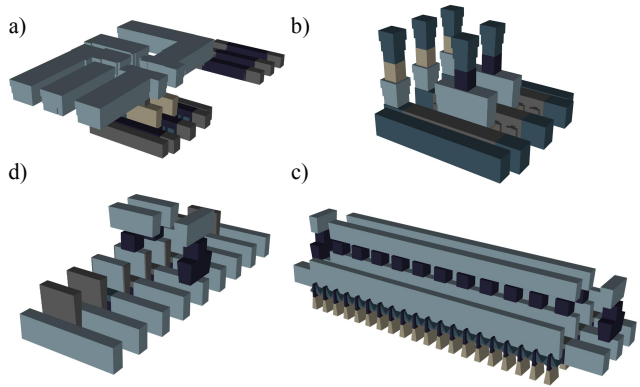}
    \caption{Test case visualizations. (a) Case 7, (b) Case 8, (c) Case 9, and (d) Case 10}
    \label{fig:case_visuals}
\end{figure}

Table~\ref{tab:testcase_description} summarizes the structural characteristics of the ten capacitance-extraction test cases, including their layer counts, conformal regions, block decompositions, and conductor counts. Figure~\ref{fig:case_visuals} provides example visualizations for Cases~7–10, illustrating the variation in geometric complexity across the dataset. These cases highlight the more intricate dielectric and conductor configurations that drive the differences observed in the runtime comparison.

\begin{table}[b]
\centering
\setlength{\tabcolsep}{3pt}
\begin{tabular}{crrrr}
\toprule
\textbf{Case} &
{\textbf{\# Layers}} &
{\textbf{\# Conformal}} &
{\textbf{\# Blocks}} &
{\textbf{\# Conductors}} \\
\midrule
1  & 58 & 1482 & 1888 & 251 \\
2  & 57 & 5838 & 14592 & 331 \\
3  & 58 & 7062 & 8383 & 1181 \\
4  &  50 &   42 &   71 &  24 \\
5  &  32 &   38 &   63 &  22 \\
6  &  32 &   46 &   75 &  26 \\
7  & 57 &  244 &  520 &  12 \\
8  & 57 &  150 &  277 &   9 \\
9  & 45 &  272 &  914 &  47 \\
10 & 42 & 1114 & 4511 &  48 \\
\bottomrule
\end{tabular}
\caption{Capacitance extraction test cases breakdown.}
\label{tab:testcase_description}
\end{table}

\section{Detailed Runtime Comparison}

The detailed runtimes in Table~\ref{tab:runtime_comparison} complement the values in Fig.~\ref{fig:runtime_comparison}. FRW-FDM and GE-CNN runtimes are added for additional context. The high FRW-FDM costs arise from repeatedly solving non-stratified dielectric regions with a finite–difference mesh, making it impractical for large-scale extraction. GE-CNN also shows notable overhead for several practical reasons: (1) its group-equivariant convolutions are custom kernels that cannot benefit from TensorRT, (2) the 24-element symmetry group multiplies the work per convolution, and (3) the GMM sampling procedure does not inherently restrict samples to cube faces, requiring boundary checks and resampling.

\begin{table}[h]
\centering
\begin{tabular}{crrrrr}
\toprule
\multirow{3}{*}{\textbf{Case}} &
\multicolumn{5}{c}{\textbf{Runtime } (s)} \\
\cmidrule(lr){2-6}
 & \multicolumn{1}{c}{\shortstack{FRW-\\FDM}}
 & \multicolumn{1}{c}{\shortstack{GE-\\CNN}}
 & \multicolumn{1}{c}{\shortstack{FRW-\\AGF}}
 & \multicolumn{1}{c}{\shortstack{Micro-\\walk}}
 & \multicolumn{1}{c}{\shortstack{DeepRW\\Cap (ours)}} \\
\midrule
1  & 2360.7 & 243.1 &  9.4 & 36.6 & 29.3 \\
2  &  603.4 &  38.6 &  2.3 &  6.4 &  9.7 \\
3  & 1633.7 & 107.2 & 13.0 & 33.1 & 14.6 \\
4  & 4323.2 & 190.1 &  8.6 & 33.5 & 27.4 \\
5  & 3404.9 &  88.5 & 11.7 & 29.9 & 22.2 \\
6  & 3889.0 &  96.4 &  8.2 & 30.2 & 25.4 \\
7  &  171.5 &  16.1 &  2.9 &  8.0 &  3.6 \\
8  &   97.7 &  10.7 &  2.0 &  7.0 &  2.2 \\
9  &  691.8 &  37.6 & 11.9 & 17.8 & 13.7 \\
10 &  395.1 &  25.7 & 11.3 & 18.2 &  9.9 \\
\bottomrule
\end{tabular}
\caption{Detailed Runtime Comparisons (seconds).}
\label{tab:runtime_comparison}
\end{table}

\section{Additional Details for Ablation Study}

This section provides detailed architectural specifications for the ablation study models.

\paragraph{MLP Baseline}
The MLP baseline flattens the $23^3$ volumetric input into 12,167 features and processes them through three hidden layers of 2048 units each with BatchNorm1d and GELU activation. The output is reshaped to a 23$\times$23 face grid. This direct volumetric-to-face mapping approach is the most parameter-heavy baseline due to the large fully connected layers.

\paragraph{3D Convolutional Baseline}
The 3D CNN uses five layers with stride-2 depth reduction (23$\rightarrow$12$\rightarrow$6$\rightarrow$3$\rightarrow$1) and channel progression 1$\rightarrow$8$\rightarrow$8$\rightarrow$16$\rightarrow$32$\rightarrow$1. Stride patterns of (2,1,1) compress the depth dimension while preserving spatial dimensions through appropriate padding. Each layer uses 3$\times$3$\times$3 kernels except the final layer which uses 1$\times$3$\times$3. Despite having relatively few parameters, the 3D operations are computationally expensive.

\paragraph{2D Convolutional Baseline}
The standard 2D CNN uses conventional convolutions with channel progression 16$\rightarrow$16$\rightarrow$8$\rightarrow$4$\rightarrow$2 and dilation pattern 1, 1, 2, 3 across four ConvBlocks. Each block consists of 3$\times$3 conv, BatchNorm2d, and GELU activation. This architecture serves as a direct comparison point for evaluating the efficiency gains of depthwise separable convolutions.

\paragraph{GE-CNN + GMM}
This approach combines group-equivariant 3D convolutions using S4 group symmetries (24 rotational transformations) with a Gaussian mixture decoder. The encoder produces 384-dimensional features (24$\times$16 channels), while the GMM decoder consists of a three-layer MLP (384$\rightarrow$500$\rightarrow$200$\rightarrow$100) that predicts parameters for 10 Gaussian components: mixing weights, 2D means, and diagonal covariance matrices. The final output is generated by evaluating the mixture density on a normalized coordinate grid.

\paragraph{Depthwise Separable Convolutions}
The depthwise separable convolutions decompose standard convolutions into depthwise (3$\times$3) and pointwise (1$\times$1) operations for parameter efficiency. Four depthwise separable blocks follow channel progression 16$\rightarrow$16$\rightarrow$8$\rightarrow$4$\rightarrow$2 with dilation pattern 1, 1, 2, 3 to expand receptive fields without increasing parameters. Each block performs depthwise conv $\rightarrow$ BatchNorm2d $\rightarrow$ GELU $\rightarrow$ pointwise conv $\rightarrow$ BatchNorm2d $\rightarrow$ GELU.

\paragraph{DS Conv + Learnable Positional Encoding}
This variant extends the base depthwise separable architecture by adding two learnable parameter channels (initialized with random normal distribution) to the input, expanding the first layer to 25 channels. 

\paragraph{DS Conv + Grid Positional Encoding}
Our best-performing architecture concatenates fixed normalized $x,y$ coordinate channels to the input volume, providing explicit spatial awareness. The coordinate grids are registered as non-trainable buffers, with the additional input channels requiring only a small expansion of the first convolutional layer.

\end{document}